\documentclass[journal]{IEEEtran}

\usepackage{cite}
\usepackage[pdftex]{graphicx}
\usepackage{amsmath}
\usepackage{amsthm,amssymb}
\usepackage{algorithmic}
\usepackage{array}
\usepackage[caption=false,font=footnotesize]{subfig}
\usepackage{fixltx2e}
\usepackage{stfloats}
\DeclareSymbolFont{AMSb}{U}{msb}{m}{n}  
\DeclareMathSymbol{\Sph}{\mathbin}{AMSb}{"53} \DeclareMathSymbol{\R}{\mathbin}{AMSb}{"52}
\DeclareMathSymbol{\T}{\mathbin}{AMSb}{"54} \DeclareMathSymbol{\Z}{\mathbin}{AMSb}{"5A}
\DeclareMathSymbol{\K}{\mathbin}{AMSb}{"4B} \DeclareMathSymbol{\N}{\mathbin}{AMSb}{"4E}
\DeclareMathSymbol{\E}{\mathbin}{AMSb}{"45}
\DeclareMathSymbol{\W}{\mathbin}{AMSb}{"57}

\usepackage[ruled,vlined]{algorithm2e}
\usepackage{yhmath}
\usepackage{mathdots}
\usepackage{MnSymbol}

\usepackage[colorlinks]{hyperref}
\usepackage[utf8]{inputenc} 
\usepackage[T1]{fontenc}    
\usepackage{hyperref}       
\usepackage{url}            
\usepackage{booktabs}       
\usepackage{amsfonts}       
\usepackage{nicefrac}       
\usepackage{microtype}      
\usepackage{verbatim} 
\usepackage{makecell}

\begin{document}

\title{Local Critic Training for Model-Parallel Learning\\of Deep Neural Networks}

\author{Hojung~Lee, 
	Cho-Jui~Hsieh,
	Jong-Seok~Lee, \textit{Senior Member}, \textit{IEEE}
	\thanks{This work was supported by Basic Science Research Program through the National Research Foundation of Korea (NRF) funded by the Korea government (MSIT) (NRF2016R1E1A1A01943283) and the Artificial Intelligence Graduate School Program (Yonsei University, 2020-0-01361). A preliminary version of this work was presented in part at the International Joint Conference on Neural Networks (IJCNN), Budapest, Hungary, July 2019 \cite{hojung19}.}%
	\thanks{H. Lee and J.-S. Lee are with the School of Integrated Technology, Yonsei University, Incheon, 21983, Korea. 
		E-mail: \{hjlee92, jong-seok.lee\}@yonsei.ac.kr}
	\thanks{C.-J. Hsieh is with Department of Computer Science, University of California at Los Angeles (UCLA), CA, USA.
		E-mail: chohsieh@cs.ucla.edu}}


\maketitle

\begin{abstract}
	In this paper, we propose a novel model-parallel learning method, called \textit{local critic training}, which trains neural networks using additional modules called \textit{local critic networks}.   
	The main network is divided into several layer groups and each layer group is updated through error gradients estimated by the corresponding local critic network. 
	We show that the proposed approach successfully decouples the update process of the layer groups for both convolutional neural networks (CNNs) and recurrent neural networks (RNNs).
	In addition, we demonstrate that the proposed method is guaranteed to converge to a critical point.
	We also show that trained networks by the proposed method can be used for structural optimization.
	Experimental results show that our method achieves satisfactory performance, reduces training time greatly, and decreases memory consumption per machine.
	Code is available at https://github.com/hjdw2/Local-critic-training.
\end{abstract}


\IEEEpeerreviewmaketitle

\section{Introduction}
 	\IEEEPARstart{R}{ecently}, deep learning has been successfully applied in many fields, including speech recognition \cite{Nassif19},\cite{Hinton12}, machine translation \cite{Singh17}, \cite{Zhang15}, image recognition \cite{Jia17}, \cite{Redmon17}, etc.
 	This achievement is mainly due to the development of large-sized neural network architectures having increased learning capabilities and the advancement of devices that can handle the huge amount of calculation for training such neural networks.
 	
	However, as the size of neural networks grows, the amounts of computation and memory consumption that a machine needs to handle increase significantly, which often becomes infeasible.
    A potential way to alleviate this issue is \textit{model-parallel learning} by exploiting multiple computing nodes simultaneously. 
    In this approach, a deep neural network is divided into several modules and then each module is distributed to a different computing node for efficient computation.
	However, the backpropagation training method that is commonly used is not suitable for this type of learning due to its sequential nature:
	The given data have to be processed through the entire network in the feedforward direction, producing an output. 
	Then, the output is compared to the target using a loss function to produce an error signal.
	The error signal is propagated in the backward direction from the output layer to the former layers to obtain the error gradient by the chain rule, based on which the weight parameters of the network are updated.
	This sequential procedure ties the whole computation process into a non-breakable unit.
	Therefore, the distributed modules in model-parallel learning cannot be trained in an efficient way using the conventional backpropagation approach.

    A few methods have been proposed for model-parallel learning.
    The method of auxiliary coordinates (MAC) in \cite{Carreira14} and the alternating direction method of multipliers (ADMM) in \cite{Taylor16} train a network by a sequence of minimization sub-steps without gradient descent steps.  
    These methods resolve the layer-wise dependency to some extent.
	However, because they were only applied to shallow fully-connected networks, it is difficult to expect that the methods work well for deep network structures such as convolutional neural networks (CNNs).
    In \cite{Jaderberg17}, a concept of predicting error gradients of layers is proposed, called the decoupled neural interface (DNI) method. 
    Prediction (instead of computation) of error gradients allows training of a certain layer before the complete backward pass till the layer.
    However, this method achieves poor performance when compared to conventional backpropagation as shown in \cite{Czarnecki17}.
    Besides, when the network becomes deeper, there are cases where learning does not converge \cite{Zhouyuan18}. 
     	
	In this paper, we propose a novel method to train neural networks in a model-parallel way by introducing auxiliary neural networks, called \textit{local critic networks}, to unlock dependencies in the update process of layers. 
	We call our method \textit{local critic training}.
	The main network is divided into several modules by the local critic networks and each local critic network delivers an estimated error gradient to the corresponding module.  
	In this way, each module has no dependency on the other modules except the corresponding local critic network, which enables model-parallel learning.	
	We show that the proposed method is applicable to both convolutional neural networks (CNNs) and recurrent neural networks (RNNs).
	Besides, we provide a theoretical analysis to demonstrate that the local critic training method converges to a critical point under certain conditions.
	In addition, by taking advantage of the fact that the outputs of the local critic networks indirectly approximate the output of the main network, we show that the trained networks can be used for structural optimization.
	
	The contribution of this paper is summarized as follows. 
	\begin{itemize}
	\item We propose the local critic training method for model-parallel training of deep neural networks including both CNNs and RNNs.
	Experimental results demonstrate that the proposed method achieves better performance of the trained networks than \cite{Jaderberg17} and faster training speed and reduced memory consumption than the conventional backpropagation training. 
	\item We mathematically prove the convergence of the proposed method. 
	\item We show that the proposed method naturally performs structural optimization. 
	In other words, the main and local critic networks trained by our method can form multiple networks having different levels of complexity, among which one can choose a compact one showing good performance.
	\end{itemize}
	
	The rest of the paper is organized as follows.
	Section \ref{sec:Related} provides a brief survey of the related work.
	Section \ref{sec:method} explains the algorithm and architecture of the proposed local critic training method.
	Section \ref{sec:converge} elaborates the convergence analysis.
	Extensive experimental results are provided in Section \ref{sec:exp}.
	Finally, conclusion is given in Section \ref{sec:conclusion}.

\section{Related Work}
\label{sec:Related}
\subsection{Efficient Learning} 
In order to alleviate the burden of training huge neural networks, plenty of methods have been studied from various perspectives, such as quantization \cite{Micikevicius18}, pruning \cite{Lym19}, knowledge distillation \cite{Yim17}, hyperparameter optimization \cite{Dong18, Dong19}, parallel learning \cite{Ben-Nun19}, etc.
Quantization is an approach using reduced precision floating-point numbers for weights, activations, and gradients. 
In \cite{Micikevicius18}, a method using half-precision floating point numbers is proposed, which reduces memory usage by half without performance degradation. Pruning is a method that removes weight parameters gradually during training to reduce the network size. 
In \cite{Lym19}, it is shown that the proposed structured pruning and reconfiguration method reduces the training time greatly. Knowledge distillation is an approach using a pre-trained teacher network for training a student network having reduced complexity. 
In \cite{Yim17}, using knowledge transfer with knowledge distillation, faster optimization and improved performance are achieved. Hyperparameter optimization is a method to find the optimal hyperparameters for learning efficiently than brute-force methods. There exist heuristic approaches using reinforcement learning to find the optimal hyperparameters during training and make the learning converge faster \cite{Dong18, Dong19}. 
Parallel learning uses multiple computing machines at the same time to reduce training time, which is surveyed in the following section in more detail.

\subsection{Parallel Learning} 
Parallel learning is categorized into two types: data parallelism and model parallelism.

Data parallelism basically replicates the same model on multiple machines and partitions the training dataset.
Then, each machine is fed with the partitioned data to perform the forward and backward passes. 
The calculated error gradients are gathered at the center node to update the model parameters using stochastic gradient descent (SGD).
Thus, the center node has to wait for all nodes to send the results before updating the model, which is called synchronous SGD \cite{Zinkevich10}.
We have another choice, in which the center node does not wait for all nodes and uses only the information available at the time, which is asynchronous SGD \cite{Recht11}.
Some examples of recent studies on data parallelism are as follows. A method of using minimal tensor swapping between CPU and GPU is proposed in \cite{Matzek18} to train large models beyond the GPU capacity. In \cite{Shallue19}, the effects of increasing the batch size during training for data parallelism are extensively investigated. In \cite{Ben-Nun19}, various types of concurrency for parallelism are analyzed in the viewpoints of stochastic optimization, network architecture, and communication mechanism.

In the model parallelism approach, the model is divided into several modules and each module is trained on a different machine.
This approach can speed up learning while reducing the computation burden per machine.

As a way of dividing a network, we can divide the network in terms of the arithmetic operations of layers, such as addition or multiplication.
In \cite{Dean12}, a framework for large-scale distributed training is proposed, which divides the computation process of a CNN into multiple machines, combining the asynchronous SGD.
Since then, various methods have been studied to partition a network with respect to batch dimension, data dimension, or channel dimension \cite{Gholami18, Jia18, Shazeer18, Jia19}. 
In addition, pipeline-based model parallel methods are proposed in \cite{Gaunt17, Narayanan19}.
In \cite{Cai20}, a method to automatically optimize parallelization strategies is proposed. 
A hybrid approach is also proposed in \cite{Pal19}, where multiple devices are used for different parts of the model in each data-parallel worker. 
These methods, like the data parallelism methods, are usually independent of the network structure and learning algorithm. 
Thus, they can be used together with other model-parallel learning methods (including the proposed method) described below.


The proposed method belongs to the approach that partitions the network in the layer dimension. 
%
In \cite{Carreira14, Taylor16}, the methods to train a network by solving an equality-constrained optimization problem are proposed, namely, MAC and ADMM, respectively. 
Because they do not need the gradient descesnt steps, they remove the dependencies in the update process of layers. 
However, these methods have proven to work only for simple networks.
In \cite{Jaderberg17}, the method using additional neural networks called decoupled neural interfaces (DNIs) is proposed.
The outputs of DNIs are the estimated error gradients of the layers in the main network for the update of the layers.
Since the error gradients are provided from the DNIs, the backward pass does not need to start from the loss function.
Thus, each layer can be updated independently in a layer-wise fashion. 
However, this method causes performance degradation compared to the backpropagation as shown in \cite{Czarnecki17}.
The idea of employing DNIs is also adopted in \cite{Czarnecki17}, where the DNIs approximate the output of the main network instead the error gradients of the layers.
However, it aims to improve the performance of the model, and does not implement parallel learning.
Furthermore, the methods in \cite{Jaderberg17, Czarnecki17} have been applied only to CNNs.
In contrast, we propose a method that improves these methods to enhance the performance and also facilitate model-parallel learning for both CNNs and RNNs.

\subsection{Structural Optimization and Anytime Prediction}
Structural optimization refers to the process of finding the optimal neural network structure to perform a given task, which has been one of the most challenging problems in neural networks \cite{kwon97,feng15,Cortes17,zoph17}.
Here, the optimality can be noted as either the least complex structure achieving the maximum possible performance or the structure having a good balance between the model complexity and performance.
The anytime prediction property means that given a certain computational budget that is not sufficient to perform a complete feedforward propagation through the trained network, an inference result can be still obtained using part of the network \cite{larsson17,huang18multi}.
CNNs and RNNs trained using the proposed local critic training algorithm are readily used for structural optimization and anytime prediction without any further training or optimization process.

\section{Proposed Approach}
\label{sec:method}
\begin{figure}[t]
	\centering
	\subfloat[CNN]{%
		\includegraphics[width=0.48\textwidth]{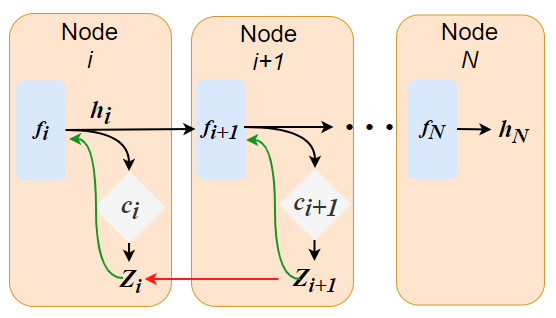}%
		\label{fig:CNN_LC}
	}
	\hfill
	\subfloat[RNN]{%
		\includegraphics[width=0.48\textwidth]{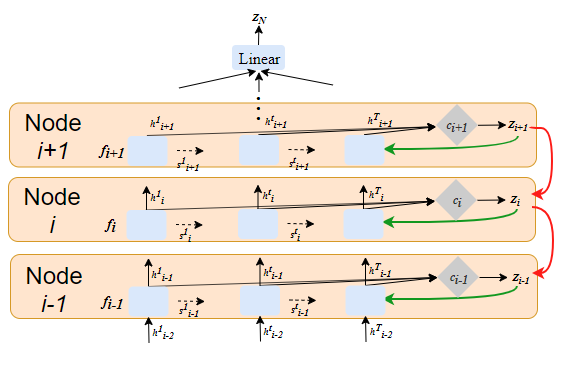}%
		\label{fig:RNN_LC}
	}
	\caption{Illustration of the proposed local critic training method. The black, green, and red arrows indicate feedforward passes, error gradient flows, and loss comparisons, respectively. Each orange box corresponds to a part that can be allocated to a separate computing node.}
	\label{fig:LC}
\end{figure}

\subsection{Local Critic Training for CNNs}
The core idea of the proposed local critic training method is to train the main network using additional networks called local critic networks.
The local critic networks are added between layers in the main network so that the main network is divided into several layer groups, and produce outputs through softmax layers in the same manner as the output of the main network. 
Each local critic network is trained in such a way that the loss of its output approximates the loss of the output of the main network. 
Then, the corresponding layer group can be trained using the output of the local critic network without the necessity of waiting for the main network to produce the output at its final layer through the complete feedforward computation.

When the main network is divided into $N$ layer groups, we denote the $i$th layer group as $f_i$ ($i=1,...,N$) in the main network and the $i$th local critic network inserted between $f_i$ and $f_{i+1}$ as $c_i$ ($i=1,...,N-1$), as shown in Figure \ref{fig:CNN_LC}. 
The output of $f_i$, denoted as $h_i$, is propagated to $c_i$, producing output $Z_i$. 
This output is compared to the target $y$ through a loss function, i.e.,
\begin{equation}
\label{eq1}
L_i = l(Z_i, y),
\end{equation}
where $l$ is the loss function such as cross-entropy or mean-squared error\footnote{More precisely, $L_i$ is a function of the weight parameters in the all layer groups before $(i+1)$th layer group, i.e., $L_i = L_i(w_1,...,w_i) = L_i(w_{1:i})$. Similarly, $L_N = L_N(w_1,...,w_N) = L_N(w_{{1:N}})$. However, for simplicity, we omit the arguments of $L_i$ in Section \ref{sec:method}.}. 
Then, the error gradient for training $f_i$ is obtained by differentiating $L_i$ with respect to $h_i$, i.e.,
\begin{equation}
\label{eq2}
\delta_{i} = \frac{\partial L_{i}}{\partial h_{i}},
\end{equation}
which can be used to update the weight parameters of $f_i$, denoted by $w_i$, via the gradient-descent rule: 
\begin{equation}
\label{eq_sgd_cnn}
w_i \leftarrow w_i - \alpha \nabla_{w_i} L_i = w_i - \alpha ~\delta_i ~\frac{\partial h_i}{\partial w_i},
\end{equation}
where $\alpha$ is a learning rate and $\nabla_{w_i}$ denotes the partial derivative with respect to $w_i$.
In order to train the weight of the layer group correctly in this way, $L_i$ has to appoximate the final output of the main network $L_N=l(h_N,y)$ so that $\delta_i$ approximates the true gradient, i.e.,
\begin{equation}
\label{eq3}
\delta_{i} \approx \frac{\partial L_{N}}{\partial h_{i}}.
\end{equation}
Thus, we can set the objective function to train $c_i$ as $l(L_i, L_{N})$, which enforces $L_i \approx L_N$. 
However, this prevents $c_i$ from being updated until $L_N$ is obtained at the output layer of the main network.
In order to alleviate such a constraint, we employ a cascaded training approach by setting the training loss for $c_i$ as
\begin{equation}
\label{eq4}
L_{c_i}=l(L_i, L_{i+1}).
\end{equation}
As a result, as learning progresses, $L_i$ can eventually approximate $L_N$.
Thus, each local critic network can be also updated to optimize the loss function given in (\ref{eq4}) once the approximated loss by the subsequent layer, $L_{i+1}$, is available.

Therefore, the loss approximation by the local critic networks and the cascaded training of the local critic networks effectively alleviate the dependencies between layers in both the forward and backward passes.
All layers in the network are dependent only on adjacent layers so that we can train the entire network in a model-parallel way by distributing the layer groups to different computing nodes as shown in Figure \ref{fig:CNN_LC}.

\subsection{Local Critic Training for RNNs}
Figure \ref{fig:RNN_LC} illustrates the architecture and training process of the proposed method for RNNs.
The $i$th layer group in the main RNN, denoted as $f_i$ ($i=1,...,N$) receives the output of the previous layer group, $h_{i-1}^t$, where $t=1,...,T$ indicates the time index, and produces its output, $h_i^t$.
The final network output $Z_N=[z_N^1, z_N^2, ..., z_N^T]^\top$ is obtained by aggregating the $N$th layer group's outputs over time (i.e., $h_N^t$) through a linear layer, where $\top$ is the transpose operation.

As in the case of CNNs, we introduce an additional local critic network $c_i$ ($i=1,...,N-1$) attached to the $i$th layer group.
It takes all the outputs of the layer group (i.e., $h_i^t$, $t=1,...,T$) as input and produces the output $Z_i=[z_i^1, z_i^2, ..., z_i^T]^\top$ that has the same dimension to the output of the main network $Z_N$.
Then, with $L_i^t = l(z_i^t,y^t)$, the loss $L_i$ of $c_i$ is obtained by
\begin{equation}
\label{eq5}
L_{i} = \sum_{t=1}^T{L_i^t},
\end{equation}
which in turn allows approximation of the error gradient:
\begin{equation}
\label{eq6}
\delta_i = \left[\frac{\partial L_i^1}{\partial h_i^1}, \frac{\partial L_i^2}{\partial h_i^2}, ..., \frac{\partial L_i^T}{\partial h_i^T} \right]^\top \approx \left[\frac{\partial L_N^1}{\partial h_N^1}, \frac{\partial L_N^2}{\partial h_N^2}, ..., \frac{\partial L_N^T}{\partial h_N^T} \right]^\top.
\end{equation}
Then, the gradient descent rule is used to update the weight parameters of $f_i$, denoted as $w_i$:
\begin{equation}
\label{eq_sgd_rnn}
w_i \leftarrow w_i - \alpha \nabla_{w_i} L_i = w_i - \alpha ~\delta_i^\top ~g_i,
\end{equation}
where $\alpha$ is a learning rate and $g_i=\left[\frac{\partial h_i^1}{\partial w_i}, \frac{\partial h_i^2}{\partial w_i}, ..., \frac{\partial h_i^T}{\partial w_i}  \right]^\top$.
In order to train the network in a model-parallel way, we use the loss function in (\ref{eq4}) to train the local critic networks.
Therefore, as in the case of CNNs, each layer group can be independently updated based on the corresponding local critic network without waiting for the complete feedforward and backward passes.

\subsection{Structure Optimization and Anytime Prediction}
\label{sec:optimization}

\begin{figure}[t]
	\centering
	\subfloat[CNN]{%
		\includegraphics[width=0.48\textwidth]{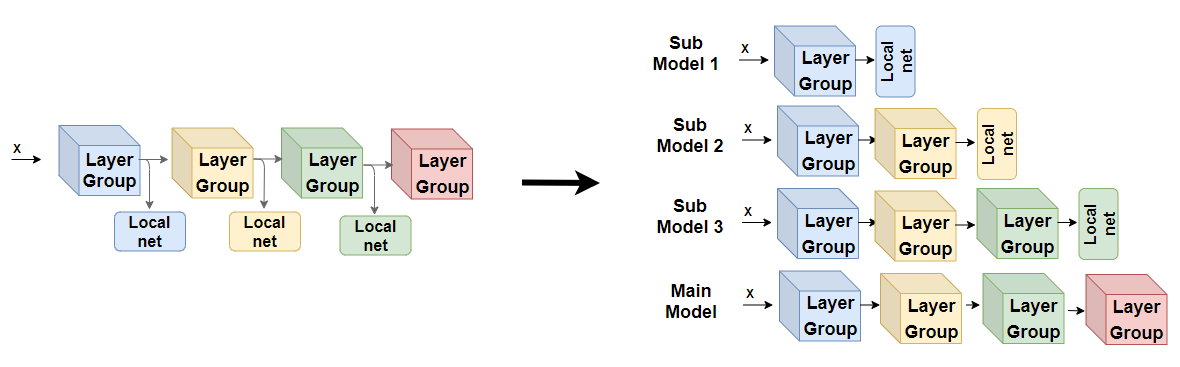}%
		\label{fig:CNN_st_op}
	}
	\hfill
	\subfloat[RNN]{%
		\includegraphics[width=0.48\textwidth]{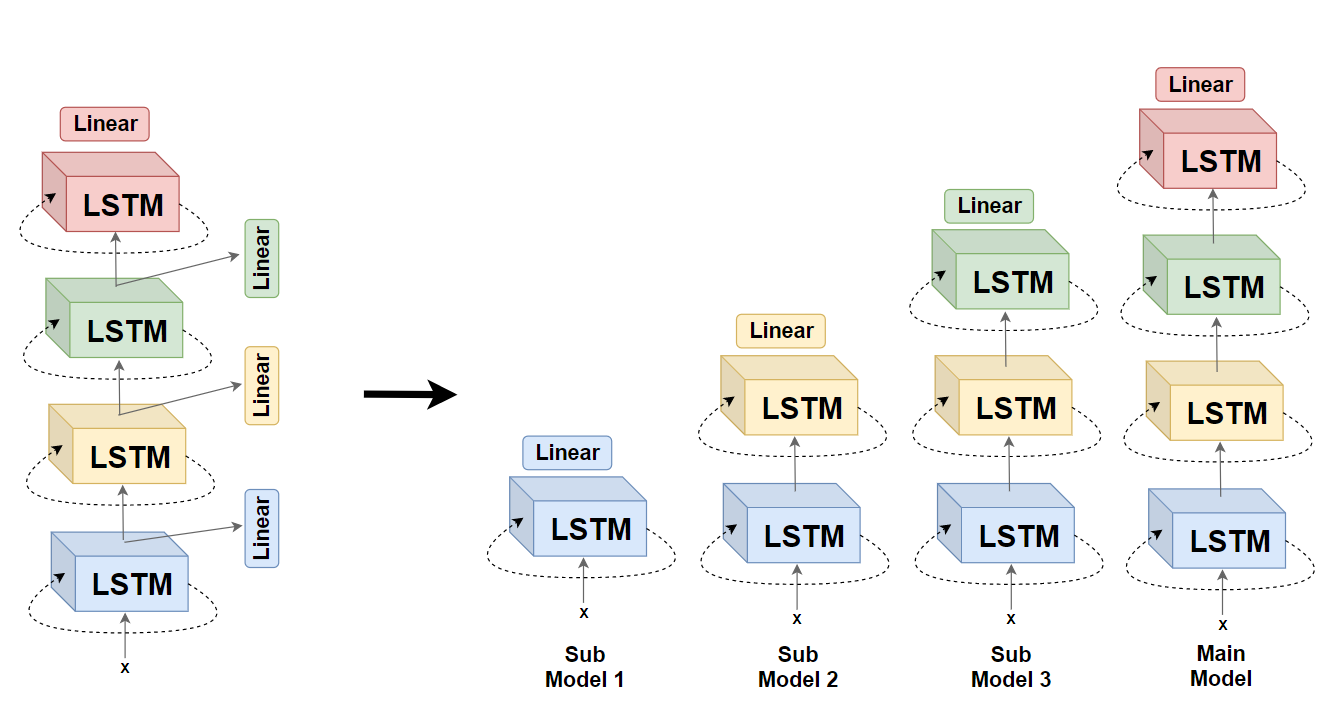}%
		\label{fig:RNN_st_op}
	}
	\caption{Sub-models obtained by the proposed training approach.}
	\label{fig:st_op}
\end{figure}

Training of $c_i$ allows it to eventually approximate the main network's output since it is trained to minimize the loss difference from the next local critic network $c_{i+1}$ by the objective (\ref{eq4}).
After the training finishes, thus, we have several sub-models that can perform the same task, as shown in Figure \ref{fig:st_op}, each of which consists of a certain number of layers and a local critic network, i.e., $f_1$ through $f_i$ and $c_i$.
Depending on the number of layers, each sub-model has different model complexity 
(in terms of the number of weight parameters and the number of floating number operations) and possibly different performance.
Among them, we can choose a suitable one by considering the trade-off between the complexity and the performance, accomplishing structural optimization for the given task.

This is also equivalent to implementing a simple anytime prediction mechanism.
In other words, if the given computational budget is small, a shallow sub-model can be used for producing output, whereas if the budget is sufficient, an output can be obtained using a deeper sub-model or even the main network, which is more reliable in general. 
Our approach is beneficial in that generic CNNs and RNNs can be made to perform anytime prediction, whereas previous approaches (e.g., \cite{larsson17,huang18multi}) rely on special network structures.

\section{Convergence Analysis}
\label{sec:converge}

This section provides a theoretical analysis for convergence of the proposed local critic training algorithm. 
By exploiting the analysis in \cite{leon19}, we prove that the algorithm is guaranteed to converge to a critical point.

Since we use the SGD method, the weight parameters of each layer group in the main network are updated according to (\ref{eq_sgd_cnn}), i.e.,
\begin{align}
w_i^{k+1} & =   w_i^k -\alpha_k \nabla_{w_i} L_i(w_{1:i}^k, \xi_k),
\end{align}
where $w_{i}^k$ is the weight parameters of the $i$th layer group at the $k$th training iteration ($k=1,...,K$), $w_{1:i}^k$ is the collection of the weight parameters in the first to $i$th layer groups at the $k$th training iteration, and $\xi_k$ is the random variable for a set of samples in the mini-batch SGD method\footnote{In Section \ref{sec:method}, we omitted the dependence of $\xi_k$ for simplicity.}.
Thus, we have the following update rule for all the weight parameters of the main network:
\begin{align}
w_{1:N}^{k+1} & =   w_{1:N}^k -\alpha_k \nabla_w L(w^k,\xi_k),
\end{align}
where
\begin{align}
\nabla_w L(w^k, \xi_k) = \begin{bmatrix}
\nabla_{w_1} L_1(w_1^k, \xi_k) \\
\vdots \\
\nabla_{w_N} L_N(w_{1:N}^k, \xi_k) \\
\end{bmatrix}.
\end{align}

First, as in \cite{Zhouyuan18}, we build a connection between the true gradient and the stochastic gradients from the local critic networks in Assumption 1. 

\textbf{Assumption 1} 
It is assumed that $w_{1:N}$ is updated in a descending direction of the true loss by the local critic training method. 
For this, the estimated stochastic gradient direction from the local critic networks, $\nabla_w L(w^k,\xi_k)$, is assumed to be a sufficient descent direction of the true loss, $\nabla L_N$. In other words, there exists a constant $\sigma > 0$ such that
\begin{align}
\label{as1}
\nabla L_N(w_{1:N}^k)^\top 
\nabla_w L(w^k, \xi_k) \geq \sigma \left\Vert \nabla L_N(w_{1:N}^k) \right\Vert_2^2.
\end{align}

This means that when each layer group is trained by the gradient from the loss function of the corresponding local critic network, the learning direction is similar to the direction of the gradient from the loss function of the main network. 
We demonstrate the validity of this assumption in Section \ref{sigma_ex} through experiments.

Next, we assume that the second moment of the stochastic gradient descent is upper bounded to restict the variance of the stochastic gradient descent. 

\textbf{Assumption 2} 
It is assumed that there exists a constant $M \geq 0 $ such that 
\begin{eqnarray}
	\E_{\xi_k} \left[ \left\Vert  \nabla_w L(w^k, \xi_k) \right\Vert_2^2 \right] \leq M.
\end{eqnarray}

In other words, the magnitude of the estimated gradient is bounded and does not diverge. This is valid when the weights are in a bounded set, which would hold in general.

Finally, we assume smoothness of the objective function, i.e., the gradient of the objective function does not change arbitrarily quickly with respect to the weight parameters.

\textbf{Assumption 3} 
It is assumed that the objective function $L_N$ is continuously differentiable and the gradient of $L_N$ is Lipschitz continuous with Lipschitz constant $\lambda > 0$, i.e., 
\begin{eqnarray}
	\left\Vert \nabla L_N(w_{1:N}) - \nabla L_N(w_{1:N}^\prime)\right\Vert_2 \; \leq \lambda \left\Vert w_{1:N} - w_{1:N}^\prime\right\Vert_2. 
\end{eqnarray}

The following theorem shows that the proposed method converges to a critical point.

\textbf{Theorem 1}
Suppose that the local critic training method is run with a learning rate sequence satisfying $\sum\limits_{k=0}^{\infty} \alpha_k = \infty$ and $\sum\limits_{k=0}^{\infty} \alpha_k^2 < \infty$.
Under Assumptions 1, 2, and 3, and if we further assume that the objective
function $L_N$ is twice differentiable, and the mapping $w_{1:N} \rightarrow \left\Vert \nabla L_N(w_{1:N}^k) \right\Vert_2^2 $ has Lipschitz continuous derivatives, then
\begin{eqnarray}
	\lim_{k \to \infty} \E \left[ \left\Vert \nabla L_N(w_{1:N}^k) \right\Vert_2^2 \right] = 0.
\end{eqnarray}
Since the expected squared norm of the gradient converges to zero, the proposed local critic training method converges to a critical point. 
The proof of the theorem is given in Appendix 1.
 
\section{Experiments}
\label{sec:exp}
In this section, we present experimental results to examine the performance of our method in various aspects, including classification accuracy, training time, memory consumption, and structural optimization for CNNs and RNNs.

For CNNs, we evaluate the method with ResNet models \cite{he16deep} (ResNet-50 and ResNet-101) on three image classification benchmark datasets: \textsf{CIFAR-10}, \textsf{CIFAR-100} \cite{Krizhevsky09}, and \textsf{ImageNet} \cite{Russakovsky15}.
We employ the SGD with a momentum of 0.9 for the main networks and the Adam optimization with a fixed learning rate of $1 \times 10^{-4}$ for the local critic networks. 
The L2 regularization is used with a fixed constant of $5 \times 10^{-4}$ for the main networks. 
For the loss functions in (\ref{eq1}) and (\ref{eq4}), we use the cross-entropy and the L1 loss, respectively, which is determined empirically. 
For \textsf{CIFAR-10} and \textsf{CIFAR-100}, the batch size is set to 128 and the maximum training iteration is set to 80,000. 
The learning rate for the main networks is initialized to 0.1 and dropped by an order of magnitude after 40,000 and 60,000 iterations.
We use one convolutional layer with the ReLU activation function and a fully-connected layer for each local critic network.
For \textsf{ImageNet}, the batch size is set to 32 and the maximum training epoch is set to 75. 
The learning rate for the main networks is initialized to 0.0125 and dropped by an order of magnitude after 23, 40, 60, and 70 epochs.
We use two convolutional layers with the ReLU activation function and a 1-channel convolutional layer for each local critic network.

For RNNs, two character-level datasets, Penn Tree Bank (\textsf{PTB}) \cite{markus93} and Hutter Wikipedia Prize (\textsf{enwik8}) \cite{Hutter12}, are used for benchmarking with long short-term memory (LSTM) units \cite{Hoch97}.
The bits per character (BPC) is used as a measure of performance.
The structures of the main network and each local critic network are a multi-layer LSTM cell and a fully-connected layer, respectively. 
As in the case of CNNs, the cross-entropy and the L1 loss are used for the loss functions in (\ref{eq5}) and (\ref{eq4}), respectively. 
The batch size is set to 128, the backpropagation through time (BPTT) length is set to 150, and the embedding size is set to 128 for both datasets.
The L2 regularization is used with a fixed constant of $1.2 \times 10^{-5}$ for the main network. 
For \textsf{PTB}, we apply the Adam optimization for 100 epochs with a learning rate that is initially set to $2.7 \times 10^{-3}$ and reduced by 0.33 at every 20 epochs. 
For \textsf{enwik8}, we use the Adam optimization with a fixed learning rate of $1.0 \times 10^{-4}$ for 50 epochs. 

The locations of the local critic networks in the main network are determined in such a way that the number of layers in each layer group is distributed as evenly as possible for all cases. 
We denote the case with $n$ local critic networks by LCT\_n$n$.

All experiments are performed using TensorFlow with GTX1080Ti graphics processing units (GPUs).
The number of employed GPUs is equal to the number of layer groups, i.e., $N$.
Each layer group and the corresponding local critic network are allocated to each GPU.


\subsection{Sufficient Descent Direction}
\label{sigma_ex}
To demonstrate the validity of the analysis in Section \ref{sec:converge}, we experimentally verify that Assumption 1 is satisfied.
For each training epoch, we calculate the gradient from the loss function of the main network, $\nabla L_N(w_{1:N}^k)$, and its squared magnitude, $\left\Vert \nabla L_N(w_{1:N}^k) \right\Vert_2^2$, and the gradient from the loss function of the local critic network, $\nabla_w L(w^k, \xi_k)$, for each layer group. Since the inequality (\ref{as1}) can be written as
\begin{align}
\frac{\nabla L_N(w_{1:N}^k)^\top \nabla_w L(w^k, \xi_k)}{\left\Vert \nabla L_N(w_{1:N}^k) \right\Vert_2^2} \geq \sigma,
\end{align}
we compute $\nabla L_N(w_{1:N}^k)^\top \nabla_w L(w^k, \xi_k)/\left\Vert \nabla L_N(w_{1:N}^k) \right\Vert_2^2$ to obtain the maximum value of $\sigma$ for each layer group and check if this value is greater than zero during training.

\begin{figure}[t]
	\centering
	\subfloat[LCT\_n1 for \textsf{CIFAR-10}\label{subfig:LC1_sigma_CIFAR10}]{%
		\includegraphics[width=0.48\columnwidth]{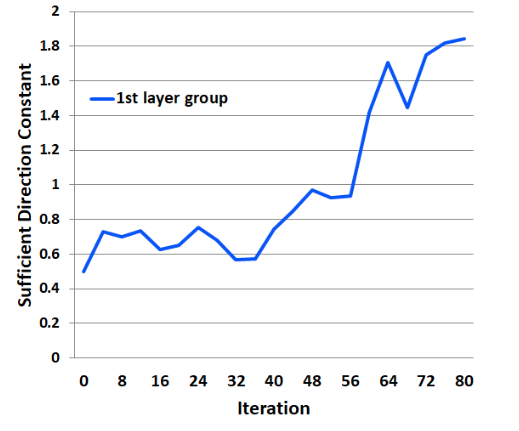}
	}
	\hfill
	\subfloat[LCT\_n3 for \textsf{CIFAR-10}\label{subfig:LC3_sigma_CIFAR10}]{%
		\includegraphics[width=0.48\columnwidth]{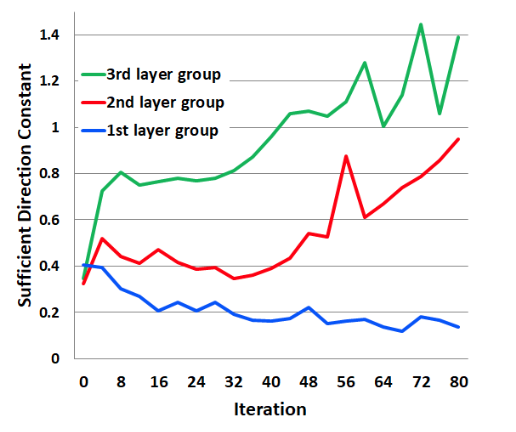}
	}
	\hfill
	\subfloat[LCT\_n1 for \textsf{CIFAR-100}\label{subfig:LC1_sigma_CIFAR100}]{%
	\includegraphics[width=0.48\columnwidth]{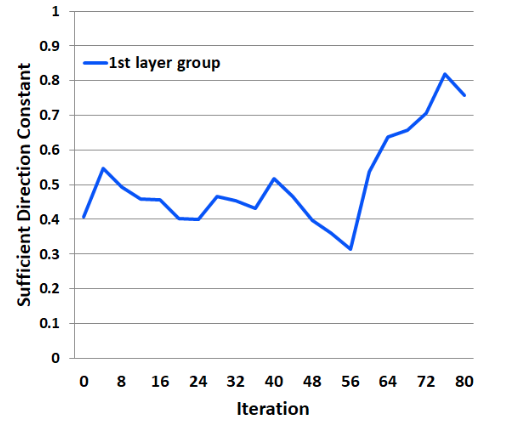}
	}
	\hfill
	\subfloat[LCT\_n3 for \textsf{CIFAR-100}\label{subfig:LC3_sigma_CIFAR100}]{%
	\includegraphics[width=0.48\columnwidth]{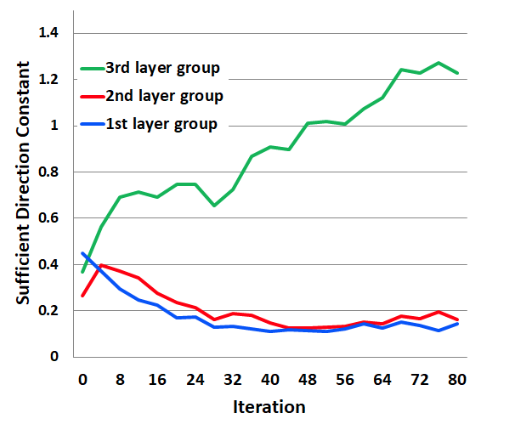}
	}
	\hfill
	\subfloat[LCT\_n1 for \textsf{ImageNet}\label{subfig:LC1_sigma_ImageNet}]{%
		\includegraphics[width=0.48\columnwidth]{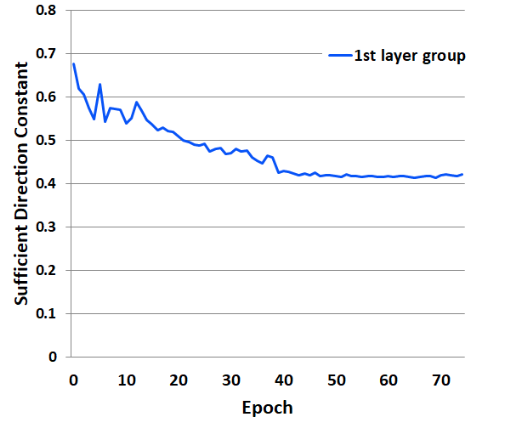}
	}
	\hfill
	\subfloat[LCT\_n3 for \textsf{ImageNet}\label{subfig:LC3_sigma_ImageNet}]{%
		\includegraphics[width=0.48\columnwidth]{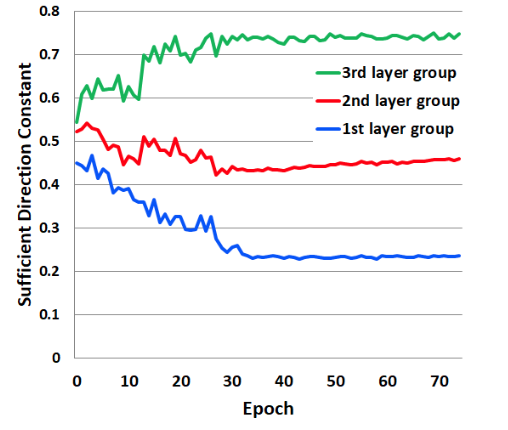}
	}
	\caption{Value of constant $\sigma$ during training for ResNet-50}
	\label{fig:CNN_sigma}
\end{figure}
\begin{figure}[t]
	\centering
	\subfloat[LCT\_n1 for \textsf{PTB}\label{subfig:PTB_LC1_sigma}]{%
		\includegraphics[width=0.48\columnwidth]{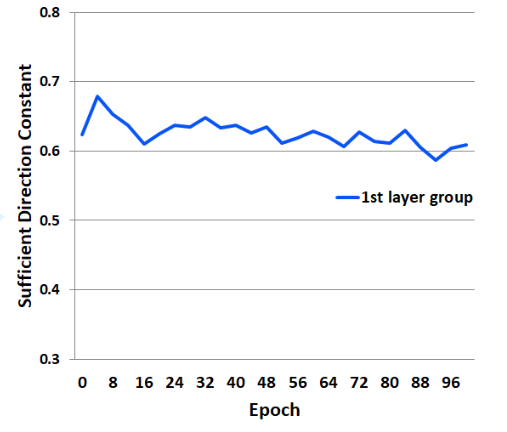}
	}
	\hfill
	\subfloat[LCT\_n3 for \textsf{PTB}\label{subfig:PTB_LC3_sigma}]{%
		\includegraphics[width=0.48\columnwidth]{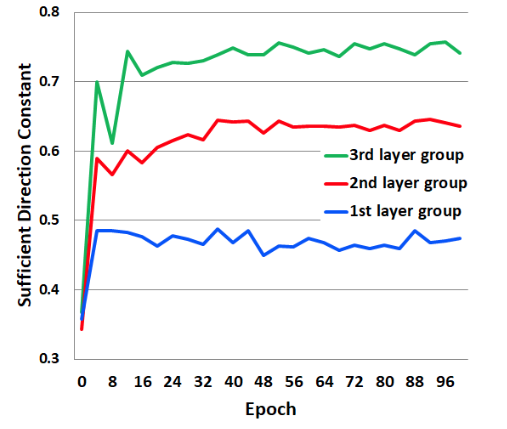}
	}
	\hfill
	\subfloat[LCT\_n1 for \textsf{enwik8}\label{subfig:enwik8_LC1_sigma}]{%
		\includegraphics[width=0.48\columnwidth]{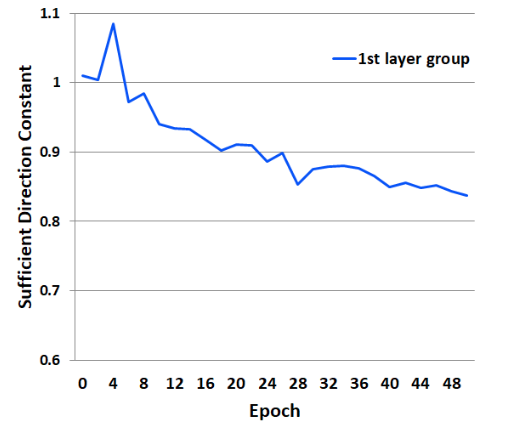}
	}
	\hfill
	\subfloat[LCT\_n3 for \textsf{enwik8}\label{subfig:enwik8_LC3_sigma}]{%
		\includegraphics[width=0.48\columnwidth]{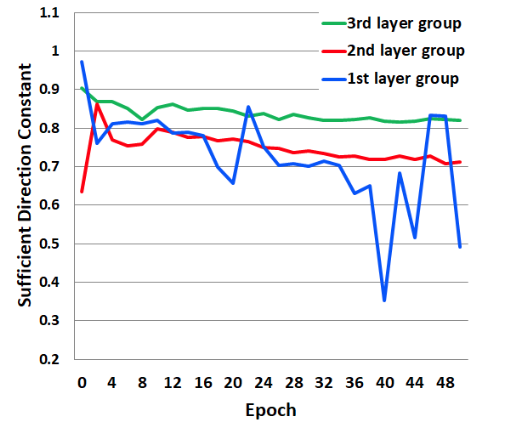}
	}
	\caption{Value of constant $\sigma$ during training for RNNs with four layers of LSTM units}
	\label{fig:RNN_sigma}
\end{figure}

Figures \ref{fig:CNN_sigma} and \ref{fig:RNN_sigma} examine the obtained maximum value of $\sigma$ in each layer group during training for the cases with one local critic network (LCT\_n1) and three local critic networks (LCT\_n3). We employ ResNet-50 for CNNs and four layers of LSTM units for RNNs. 
We omit the result of the last layer group in all cases since the last layer group is trained by the regular backpropagation and thus $\sigma$ is always one.

The figures confirm that $\sigma$ is larger than 0 at all times during training for all cases of CNNs and RNNs. Therefore, Assumption 1 is satisfied and the convergence analysis in Section \ref{sec:converge} is valid.

\subsection{Performance Evaluation}
\subsubsection{CNN}
\label{subsubsec:CNN_perf}

\begin{table*}[t]
	\renewcommand{\arraystretch}{1.6}
	\caption{Test accuracy (\%) of backpropagation (BP), DNI, and the proposed local critic training method (LCT).}
	\label{table:CNN_performance}
	\centering
	\begin{tabular}{lccccccccccr}
		\hline
		ResNet-50         & BP & DNI\_n1 & DNI\_n2 & DNI\_n3 & LCT\_n1 & LCT\_n2 & LCT\_n3\\
		\hline
		\textsf{CIFAR-10}  & 93.79 & 80.07 & 74.59 & 71.24 & 91.41 & 92.64 & 93.32\\
		\textsf{CIFAR-100} & 76.16 & 49.44 & 43.39 & 43.31 & 73.65 & 71.37 & 70.90\\
		\textsf{ImageNet}  \makecell{Top-1 \\ Top-5} & \makecell{72.09 \\ 90.51} & - & - & - &  \makecell{67.80 \\ 88.24} & \makecell{67.67 \\ 88.07} & \makecell{65.81 \\ 86.79}\\
		\hline
		ResNet-101 & BP & DNI\_n1 & DNI\_n2 & DNI\_n3 & LCT\_n1 & LCT\_n2 & LCT\_n3 \\
		\hline
		\textsf{CIFAR-10}  & 93.77 & 75.70 & 66.53 & 65.46 & 93.25 & 93.10 & 93.34 \\
		\textsf{CIFAR-100} & 76.76 & 43.94 & 35.37 & 33.62 & 71.82 & 72.14 & 72.10 \\
		\textsf{ImageNet}  \makecell{Top-1 \\ Top-5} &  \makecell{73.38 \\ 91.37} & - & - & - & \makecell{72.11 \\ 90.66} & \makecell{68.20 \\ 88.34} & \makecell{67.05 \\ 87.80}\\
		\hline
	\end{tabular}
\end{table*}

\begin{table}[t]
	\renewcommand{\arraystretch}{1.6}
	\caption{Test accuracy (\%) of backpropagation (BP) and the proposed local critic training method (LCT) for ResNet-182.}
	\label{table:CNN_large}
	\centering
	\begin{tabular}{lccccccccccr}
		\hline
		 & BP & LCT\_n1 & LCT\_n2 & LCT\_n3\\
		\hline
		\textsf{CIFAR-10}  & 94.80 & 92.89 & 94.10 & 93.00 \\
		\textsf{CIFAR-100} & 77.46 & 73.29 & 72.50 & 72.74 \\
		\textsf{ImageNet}  \makecell{Top-1 \\ Top-5} & \makecell{75.42 \\ 92.33} & \makecell{74.13 \\ 91.63} & \makecell{70.62 \\ 89.82} & \makecell{66.46 \\ 87.43} \\
		\hline
	\end{tabular}
\end{table}

\begin{table}[t]
	\renewcommand{\arraystretch}{2.0}
	\caption{Test accuracy (\%) of backpropagation (BP), ADMM, DNI, and the proposed local critic training method (LCT) for a multilayer perceptron having 300, 150, and 10 units.}
	\label{table:CNN_shallow}
	\centering
	\begin{tabular}{lccccccccccr}
		\hline
		 & BP & ADMM & DNI\_n1 & LCT\_n1\\
		\hline
		\textsf{MNIST}  & 98.05 & 82.53 & 95.99 & 98.09 \\
		\textsf{CIFAR-10} & 61.22 & 19.29 & 35.75 & 61.28\\
		\hline
	\end{tabular}
\end{table}

\begin{table}[t]
	\renewcommand{\arraystretch}{1.6}
	\caption{Test accuracy (\%) of backpropagation with joint losses obtained from the same network structure to LCT\_n3.}
	\label{table:BP_multi_loss}
	\centering
	\begin{tabular}{lcccr}
		\hline
		& ResNet-50 & ResNet-101 \\
		\hline
		\textsf{CIFAR-10}  & 94.80 & 94.52 \\
		\textsf{CIFAR-100} & 76.39 & 75.37 \\
		\textsf{ImageNet}  \makecell{Top-1 \\ Top-5} & \makecell{71.86 \\ 90.32} & \makecell{72.40 \\ 90.66} \\
		\hline
	\end{tabular}
\end{table}

We evaluate the classification performance of the proposed local critic training method with different numbers of local critic networks. 
We also compare the results of the regular backpropagation and DNI method \cite{Jaderberg17} as shown in Table \ref{table:CNN_performance}.
For the DNI method, the layers of the main network are grouped in the same way to our method, and each DNI is implemented with three convolutional layers, which shows the best performance in our experiment.

For \textsf{CIFAR-10}, the proposed method achieves comparable performance to the regular backpropagation.
The performance varies with the number of local critic networks (and the number of layer groups, accordingly), but the test accuracy of the proposed method with four layer groups is almost the same to that of the backpropagation (93.79\% vs. 93.32\% with ResNet-50, and 93.77\% vs. 93.34\% with ResNet-101).
When compared to the DNI method, our method achieves much better performance in all cases.
For \textsf{CIFAR-100}, the test accuracy of our method is slightly lower when compared to that of the backpropagation.
However, it is significantly higher than that of the DNI approach.
Similarly, for \textsf{ImageNet}, the proposed method shows only slightly lowered accuracy compared to the backpropagation, whereas DNI fails to converge in all cases.
These results show that our method can train the large-sized CNNs in a model-parallel way at the cost of slight accuracy drops and significantly outperform the existing DNI method.

Using more local critic networks allows us to exploit more computing nodes simultaneously for faster training (Section \ref{sec:complexity}), but intuitively, it causes larger accuracy drop because of more approximation stages. 
Thus, the classification accuracy and the number of local critic networks are in a tradeoff relationship, which is observed in the results for \textsf{ImageNet}.
Such a relationship is not quite clear for \textsf{CIFAR-10} and \textsf{CIFAR-100}, which is probably because those datasets are not significantly challenging and thus gradient estimation using local critic networks does not impose significant difficulty in training.

In order to validate that our method works for networks even larger than ResNet-101, we employ ResNet-182, which is the largest size that our single GPU machine (GTX1080Ti with 11GB memory) can afford. It has ten more 3-layer bottleneck blocks than ResNet-152 on the feature map of 28$\times$28 (i.e., conv4) \cite{he16deep}. Table \ref{table:CNN_large} demonstrates that our method performs well on this large-sized network, showing similar trends to those in Table \ref{table:CNN_performance}. 

In contrast, we also examine if the proposed method can achieve satisfactory performance for shallow networks compared to the existing methods including ADMM \cite{Taylor16} and DNI. For this, we use a multilayer perceptron consisting of three dense layers having 300, 150, and 10 units, respectively, with the ReLU activation function. The results for \textsf{MNIST} and \textsf{CIFAR-10} are shown in Table \ref{table:CNN_shallow}. Unlike ADMM and DNI showing degraded performance, our method achieves almost the same performance to backpropagation. 

As shown in \cite{Teerapittayanon16}, using extra layers as early exits (like our network structure) and combining the losses of the exits for backpropagation may achieve higher performance than the original network. Thus, we evaluate the performance of backpropagation using the joint losses as in \cite{Teerapittayanon16} with the same network structure used for LCT\_n3. However, the results presented in Table \ref{table:BP_multi_loss} show that it is not clear whether using joint losses improves performance than the original backpropagation. For \textsf{CIFAR-10}, the performance of backpropagation using joint losses is slightly improved than the original backpropagation, but for \textsf{CIFAR-100} and \textsf{ImageNet}, the performance is rather reduced. Since the structures of our local critic networks are extremely simple to reduce the computational burden, the losses of the early exits seem to have a negative effect on the whole learning for complex data.


\subsubsection{RNN}
\begin{table}[t]
	\renewcommand{\arraystretch}{2.2}
	\caption{Test BPC of the backpropagation (BP) and the proposed local critic training method (LCT) with varying the number of layers in the main network. The number of LSTM units per layer is also shown in each case.}
	\label{table:RNN_PTB}
	\centering
	\begin{tabular}{lccccccr}
		\hline		
	     & \makecell{$\#$ of layers \\ ($\#$ of LSTM units)} & \makecell{2 \\ (950)} & \makecell{3 \\ (750)} & \makecell{4 \\ (600)} \\
		\hline
		\textsf{PTB}    & \makecell{BP \\ LCT} & \makecell{1.275 \\ 1.276} & \makecell{1.288 \\ 1.265} & \makecell{1.275 \\ 1.269} \\
		\hline
		\textsf{enwik8} & \makecell{BP \\ LCT} & \makecell{1.476 \\ 1.469} & \makecell{1.509 \\ 1.492} & \makecell{1.618 \\ 1.541} \\
		\hline
	\end{tabular}
\end{table}

We evaluate the performance of the proposed local critic training method in comparison to the conventional backpropagation with respect to the number of layers in the main network.
In this experiment, we deploy local critic networks between every layer pair in the main network.
The number of LSTM units in each layer of the main network is determined by the memory limitation of the used GPU for backpropagation.

The results are shown in Table \ref{table:RNN_PTB}. The performance of the local critic training method is similar to or, in most cases, even better than that of the backpropagation, which demonstrates that using the estimated error gradients is effective particularly for training RNNs. Therefore, we can conclude that the proposed method can unlock the layer-wise dependencies without performance degradation over a wide range of the number of RNN layers.

\subsection{Complexity}
\label{sec:complexity}
We evaluate the computation and memory complexities of the proposed method. 
\subsubsection{Training Time}
\begin{figure*}[t]
	\subfloat[ResNet-50 for \textsf{CIFAR-10}\label{subfig:time_50_CIFAR10}]{%
		\includegraphics[width=0.98\columnwidth]{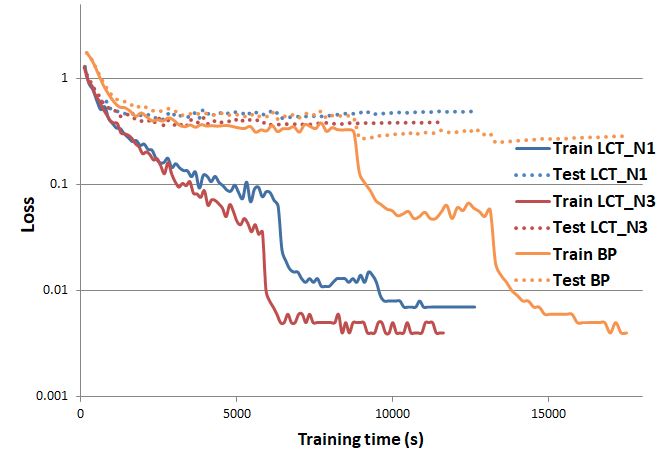}
	}
	\subfloat[ResNet-101 for \textsf{CIFAR-10}\label{subfig:time_101_CIFAR10}]{%
		\includegraphics[width=0.98\columnwidth]{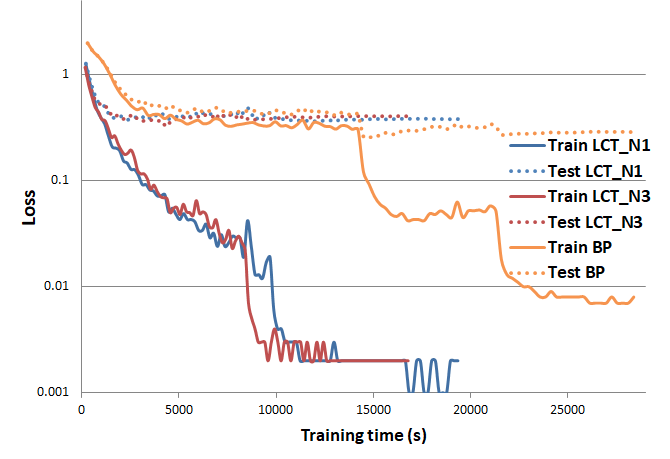}
	}
	\hfill
	\subfloat[ResNet-50 for \textsf{CIFAR-100}\label{subfig:time_50_CIFAR100}]{%
		\includegraphics[width=0.98\columnwidth]{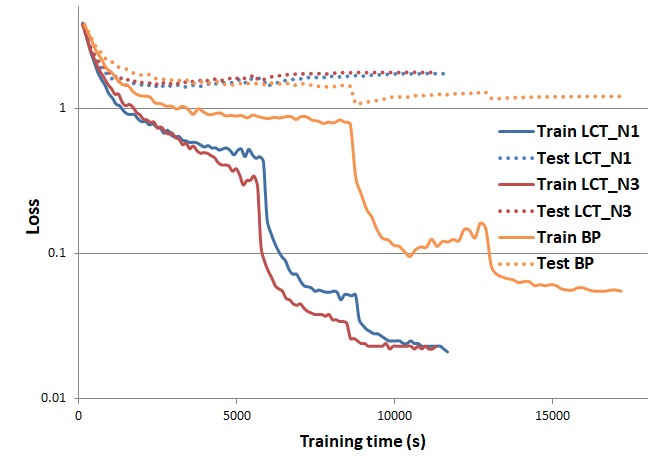}
	}
	\subfloat[ResNet-101 for \textsf{CIFAR-100}\label{subfig:time_101_CIFAR100}]{%
		\includegraphics[width=0.98\columnwidth]{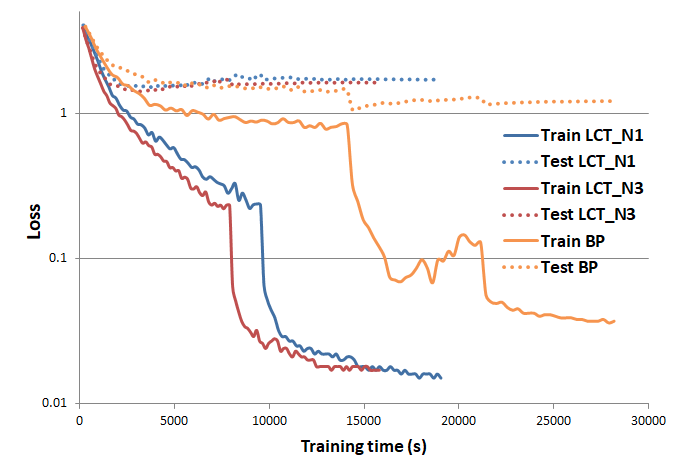}
	}
	\caption{Training and test losses with respect to the elapsed time for CNNs}
	\label{fig:CNN_time}
\end{figure*}

\begin{figure}[t]
	\includegraphics[width=0.98\columnwidth]{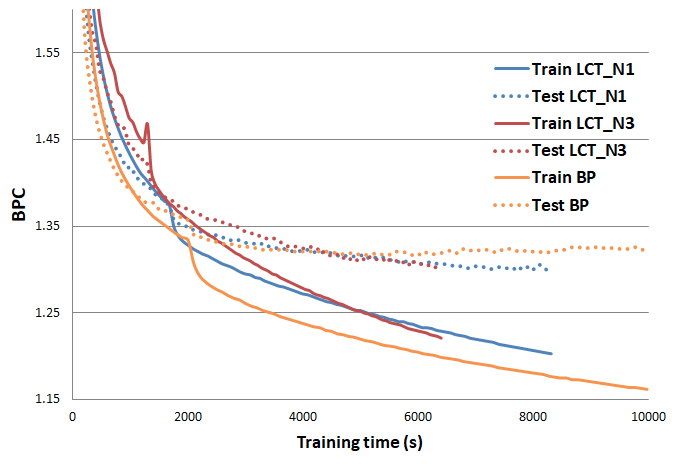}
	\caption{Training and test losses with respect to the elapsed time for the LSTM networks having four layers}
	\label{fig:RNN_time}
\end{figure}

We compare the training time of LCT\_n1, LCT\_n3, and the backpropagation for CNNs and RNNs until the maximum epoch (or iteration) reaches.
In Figure \ref{fig:CNN_time}, we show the loss with respect to the training time of ResNet-50 and ResNet-101 for \textsf{CIFAR-10} and \textsf{CIFAR-100}. 
In all cases, the training time of LCT\_n3 is the shortest, followed by that of LCT\_n1.
For \textsf{CIFAR-10}, training time is reduced by 27.8\% for LCT\_n1 and 33.6\% for LCT\_n3 in comparison to the backpropagation with ResNet-50.
With ResNet-101, the relative training time reduction is 31.9\% for LCT\_n1 and 40.9\% for LCT\_n3.
For \textsf{CIFAR-100}, the amounts of training time reduction are 32.0\% for LCT\_n1 and 34.2\% for LCT\_n3 with ResNet-50, and 32.4\% for LCT\_n1 and 43.9\% for LCT\_n3 with ResNet-101.
Figure \ref{fig:RNN_time} shows the BPC of LSTM networks having four layers with respect to the training time for \textsf{PTB}.
The training time is reduced by 16.7\% for LCT\_n1 and 35.8\% for LCT\_n3 for \textsf{PTB} in comparison to backpropagation.
These results demonstrate that the proposed method can implement efficient model-parallel training.

The training time of the proposed method includes the time for communication between different computing nodes. We investigate the relative amount of such communication time. Since it is difficult to directly measure it during training, we measure the time required for a variable of the same size with the data to be transmitted (i.e., the outputs of the layer groups for the feedforward pass and the losses from the local critic networks for the backward pass). For \textsf{CIFAR-10}, the time for communication is 3.79\% of the total training time for LCT\_n1 and 8.67\% for LCT\_n3 with ResNet-50, and 1.47\% for LCT\_n1 and 3.58\% for LCT\_n3 with ResNet-101.
Therefore, although the communication time increases as the network is divided more, its proportion to the total learning time decreases as the network becomes larger. Overall, we can say that the communication time is relatively short or even negligible. 

\begin{figure}[t]
	\includegraphics[width=0.98\columnwidth]{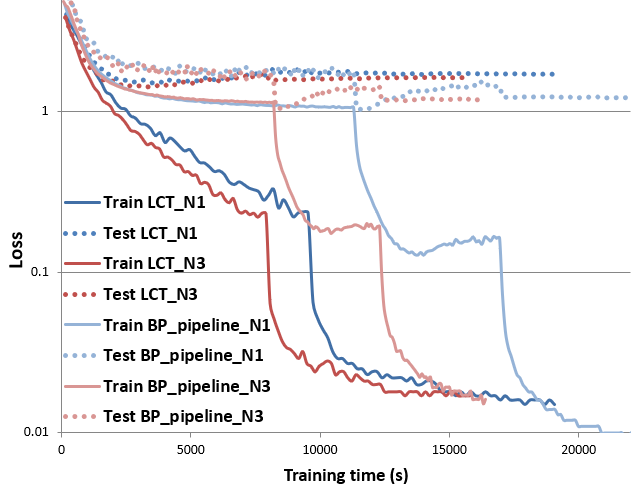}
	\caption{Training and test losses of the pipelined backpropagation and the proposed method with respect to the elapsed time with ResNet-101 for \textsf{CIFAR-100}}
	\label{fig:CNN_time_pipe}
\end{figure}

In addition, we compare our method with a simple pipelined version of backpropagation. In other words, after splitting the network into several layer groups that are allocated to different computing nodes, each node can concurrently run the feedforward pass by delivering the output to the next node and then immediately receiving a new input from the previous node. Thus, it is a basic model-parallel learning strategy that has no performance difference from the original backpropagation and can reduce the learning time. Note that, since this method is implemented in Pytorch, its result cannot be directly compared with the results shown above. In order to enable comparison between these results and the results of our method implemented in Tensorflow, we normalize the training time of the pipelined backpropagation with that of the original backpropagation. 
We compare the training time of the pipelined backpropagation using 2 GPUs and 4 GPUs, which correspond to LCT\_n1 and LCT\_n3, respectively. As shown in Figure \ref{fig:CNN_time_pipe}, for \textsf{CIFAR-100}, the training time of our method is shorter by 15.7\% for LCT\_n1 and 3.5\% for LCT\_n3 than that of the corresponding pipeline backpropagation for ResNet-101. 
Therefore, our method is more effective even when compared to a pipelined version of backpropagation.
Moreover, it would be also possible to combine this pipelining method into our method for further improvement of our method.

\subsubsection{Memory Consumption}
By splitting the computation for training among multiple computing nodes, the proposed method can additionally reduce the amount of memory usage per computing node. We compare the memory consumption per GPU by our method and the backpropagation for \textsf{ImageNet} and \textsf{PTB} in Table \ref{table:memory}. It is shown that even if the total amount of memory consumption increases, we can alleviate the burden of the memory consumption per GPU. 
Additionally, when compared to the DNI method for CIFAR-100, there is almost no difference between our method and the DNI method; the memory usage of our method is 4659, 5461, 5463, and 5459 MiB for each GPU, respectively, and that of DNI is 4693, 5453, 4943, and 5459 MiB for each GPU, respectively. 


Note that the layer grouping and the location of the local critic networks were not determined by considering the memory usage. 
If the memory constraints are significant, the layer grouping can be designed in a more memory-efficient way.

\begin{table}[t]
	\renewcommand{\arraystretch}{1.6}
	\caption{Memory consumption (MiB) per GPU by backpropagation (BP) and the proposed local critic training method (LCT\_n3) for the \textsf{ImageNet} and \textsf{PTB} datasets.}
	\label{table:memory}
	\centering
	\begin{tabular}{lccccccr}
		\hline
		ResNet-50 (\textsf{ImageNet}) & GPU 1 & GPU 2 & GPU 3 & GPU 4 \\
		\hline
		BP     & 8807 & -    & -    & -    \\
		LCT\_n3 & 3207 & 3077 & 7141 & 7171 \\	
		\hline
		ResNet-101 (\textsf{ImageNet}) & GPU 1 & GPU 2 & GPU 3 & GPU 4 \\
		\hline
		BP     & 8789 & -    & -    & -    \\
		LCT\_n3 & 4663 & 7141 & 5863 & 5127 \\		
		\hline
		4-layer LSTM (\textsf{PTB}) & GPU 1 & GPU 2 & GPU 3 & GPU 4 \\
		\hline
		BP     & 8695 & -    & -    & -    \\
		LCT\_n3 & 4573 & 2441 & 2441 & 4489 \\	
		\hline
	\end{tabular}
\end{table}

\subsection{Structural Optimization and Anytime Prediction}
\subsubsection{CNN}
As depicted in Figure \ref{fig:st_op}, we obtain several trained sub-models in addition to the trained main network as a result of application of the local critic training method.
Table \ref{table:CNN_st_op} shows their performance, and Table \ref{table:CNN_param} analyzes their complexity in terms of computational complexity (the number of floating-point operations (FLOPs)) required for one feedforward pass and memory complexity (the number of weight parameters) for the case using three local critic networks (LCT\_n3). 
In Table \ref{table:CNN_param}, the complexity of each local critic network itself with the percentage over the total complexity of the corresponding sub-model is also shown.

As expected, the larger the network is, the higher the classification accuracy is, which is reasonable because a larger network has a greater capability of learning the data. 
However, the largest network (i.e., the main network) is not necessarily the optimal structure when both the accuracy and model complexity are considered. 
For \textsf{CIFAR-10}, Sub\_3 of ResNet-50 and Sub\_2 of ResNet-100 show almost the same accuracy to the corresponding main networks, each of which reduces the complexity by about 59\% (47.55 to 19.53 million FLOPs and 23.82 to 9.79 million parameters) and 72\% (85.17 to 23.99 million FLOPs and 42.68 to 12.03 million parameters), respectively. Interestingly, these sub-models have roughly similar complexities, showing that regardless of the starting network (ResNet-50 or ResNet-100), the structural optimization results could be similar. 
For \textsf{CIFAR-100}, the same result of structural optimization can be obtained with minor accuracy loss (0.93 to 1.61\%). For a more challenging dataset, i.e., \textsf{ImageNet}, reduction of the complexities via choosing the best sub-model (i.e., Sub\_3) is obtained at the cost of slight accuracy loss.

Choosing smaller sub-models results in larger accuracy loss but more significant complexity reduction. 
Therefore, when there exist resource limitations in a target application, the model can be chosen among the sub-models and main network by considering the trade-off between the performance and the complexity.

Anytime prediction can be implemented in a similar way.
When a computational budget is given, a sub-model satisfying the budget can be chosen to produce the output.
As shown in Table \ref{table:CNN_st_op}, the more budget we have, the more accurate the prediction is.


\begin{table}[t]
	\renewcommand{\arraystretch}{1.6}
	\caption{Test accuracy (\%) of the main model and the sub-models for CNNs (LCT\_n3).}
	\label{table:CNN_st_op}
	\centering
	\begin{tabular}{lccccccr}
		\hline
		ResNet-50 & Sub\_1 & Sub\_2 & Sub\_3 & Main\\
		\hline
		\textsf{CIFAR-10}  & 86.78 & 91.40 & 93.24 & 93.32\\
		\textsf{CIFAR-100} & 60.03 & 66.71 & 69.29 & 70.90\\
		\textsf{ImageNet}  \makecell{Top-1 \\ Top-5} & \makecell{36.59 \\ 62.18} & \makecell{49.98 \\ 75.08} & \makecell{60.73 \\ 83.30} 
		& \makecell{65.81 \\ 86.79}\\		
		\hline
		ResNet-101 & Sub\_1 & Sub\_2 & Sub\_3 & Main\\
		\hline
		\textsf{CIFAR-10}  & 91.68 & 93.35 & 93.33 & 93.34\\
		\textsf{CIFAR-100} & 67.22 & 71.17 & 71.59 & 72.10\\
		\textsf{ImageNet}  \makecell{Top-1 \\ Top-5} & \makecell{49.44 \\ 75.11} & \makecell{62.68 \\ 84.80} & \makecell{63.69 \\ 85.44}
		& \makecell{67.05 \\ 87.80} \\			
		\hline
	\end{tabular}
\end{table}

\begin{table*}[t]
	\renewcommand{\arraystretch}{1.6}
	\caption{FLOPs required for a feedforward pass and numbers of model parameters in the sub-models and main model for CNNs (LCT\_n3). The complexity of each local critic network itself with the percentage over the total complexity is shown in parentheses.}
	\label{table:CNN_param}
	\centering
	\makebox[\textwidth][c]{
		\begin{tabular}{c|cc|cc|cc}
			\hline
			 & \multicolumn{2}{c|}{\textsf{CIFAR-10}} & \multicolumn{2}{c|}{\textsf{CIFAR-100}} & \multicolumn{2}{c}{\textsf{ImageNet}}\\
			\hline
			ResNet-50  & FLOP (M) & $\#$ of parameters (M) & FLOP (M) & $\#$ of parameters (M) & FLOP (M) & $\#$ of parameters (M) \\
			\hline
			Sub\_1 & 3.64 (3.21, 88\%) & 1.82 (1.61, 88\%) & 27.24 (26.80, 98\%) & 13.62 (13.40, 98\%) & 0.95 (0.50, 53\%) & 0.47 (0.25, 53\%)\\
			Sub\_2 & 4.71 (1.84, 39\%) & 2.35 (0.92, 39\%) & 10.61 (7.73, 73\%) & 5.30 (3.87, 73\%) & 3.68 (0.79, 21\%) & 1.84 (0.40, 22\%) \\
			Sub\_3 & 19.61 (2.52, 13\%) & 9.79 (1.26, 13\%) & 21.08 (4.00, 19\%) & 10.53 (2.00, 19\%) & 18.48 (1.38, 7\%) & 9.23 (0.69, 7\%) \\
			Main   & 47.69 & 23.82 & 53.58 & 26.77 & 51.15  & 25.55 \\
			\hline
			ResNet-101  & FLOP (M) & $\#$ of parameters (M) & FLOP (M) & $\#$ of parameters (M) & FLOP (M) & $\#$ of parameters (M) \\
			\hline
			Sub\_1 & 4.71 (1.84, 39\%) & 2.35 (0.92, 39\%)& 10.61 (7.73, 73\%)& 5.30 (3.87, 73\%)& 3.68 (0.79, 21\%)& 1.84 (0.40, 22\%)\\
			Sub\_2 & 24.08 (2.52, 10\%)& 12.03 (1.26, 10\%)& 25.56 (4.00, 16\%)& 12.76 (2.00, 16\%)& 25.19 (1.38, 5\%)& 12.58 (0.69, 5\%)\\
			Sub\_3 & 44.08 (2.52, 6\%)& 22.01 (1.26, 6\%)& 45.55 (4.00, 9\%)& 22.75 (2.00, 9\%)& 47.43 (1.38, 3\%)& 23.68 (0.69, 3\%)\\
			Main   & 85.44 & 42.68 & 91.34 & 45.62 & 93.38 & 46.64 \\
			\hline
		\end{tabular}
	}
\end{table*}

\subsubsection{RNN}

\begin{table}[t]
	\renewcommand{\arraystretch}{1.6}
	\caption{Test BPC of the main model and the sub-models for 4-layer LSTM networks (LCT\_n3).}
	\label{table:RNN_st_op}
	\centering
	\begin{tabular}{lcccccr}
		\hline
		 & Sub\_1 & Sub\_2 & Sub\_3 & Main \\		
		\hline
		\textsf{PTB}  & 1.329 & 1.281 & 1.272 & 1.269  \\
		\textsf{enwik8} & 1.691 & 1.585 & 1.551 & 1.541  \\	
		\hline
	\end{tabular}
\end{table}

We show the performance (BPC) of the sub-models and the main model of RNNs in Table \ref{table:RNN_st_op} and their computational complexity for one feedforward pass and the number of weight parameters in Table \ref{table:RNN_param} for the case using three local critic networks (LCT\_n3). The complexity of each local critic network itself with the percentage over the total complexity of the corresponding sub-model is also shown.

As in the CNN cases, an optimal network structure can be chosen by considering the trade-off relationship between the performance and complexity.
The largest sub-models (Sub\_3) achieve almost the same performance to that of the main model (1.272 vs. 1.269 for \textsf{PTB}, and 1.551 vs. 1.541 for \textsf{enwik8}), while they can reduce the computational and memory complexities by about 28\% (401 to 290 billion FLOPs and 10.43 to 7.55 million parameters).
The performance of the smaller sub-models is slightly worse than that of the main model, but the complexity reduction is more significant (reductions by about 55\% and 83\% with Sub\_2 and Sub\_1, respectively). Anytime prediction can be also performed using the sub-models requiring lower complexities than the main model.



\begin{table}[t]
	\renewcommand{\arraystretch}{1.6}
	\caption{FLOPs required for a feedforward pass and the numbers of model parameters in the sub-models and main model for 4-layer LSTM networks (LCT\_n3). The complexity of each local critic network itself with the percentage over the total complexity is shown in parentheses.}
	\label{table:RNN_param}
	\centering
	\begin{tabular}{lcccr}
		\hline
		 & FLOP (B) & $\#$ of parameters (M) \\	
		\hline		
		Sub\_1 & 68 (1.15, 1.7\%) & 1.79 (0.03, 1.7\%) \\
		Sub\_2 & 179 (1.15, 0.6\%) & 4.67 (0.03, 0.6\%) \\
		Sub\_3 & 290 (1.15, 0.4\%) & 7.55 (0.03, 0.4\%) \\
		Main & 401 & 10.43 \\	
		\hline
	\end{tabular}
\end{table}

\section{Conclusion}
\label{sec:conclusion}

In this paper, we proposed the local critic training method for model-parallel training of CNNs and RNNs. 
The mathematical analysis showed the convergence of the proposed method.
Through the experiments, we confirmed the effectiveness of the method, including the satisfactory classification performance, faster training speed, and lower memory consumption per GPU. 
It was also shown that structural optimization and anytime prediction can be achieved using the models trained by the proposed method.

\ifCLASSOPTIONcaptionsoff
\newpage
\fi




\bibliography{Reference}
\bibliographystyle{IEEEtran}
\onecolumn

\appendices
\section{Proof of Theorem 1}
Let us begin with rewriting $L_N(w_{1:N}^{k+1})$ as 
\begin{align}   
L_N(w_{1:N}^{k+1}) & = L_N(w_{1:N}^k) + \int_{0}^{1} \frac{\partial L_N(w_{1:N}^k + \tau(w_{1:N}^{k+1}-w_{1:N}^{k}))}{\partial \tau} \, d\tau \\
&=  L_N(w_{1:N}^k) + \int_{0}^{1} \nabla L_N(w_{1:N}^k + \tau(w_{1:N}^{k+1}-w_{1:N}^{k}))^\top (w_{1:N}^{k+1}-w_{1:N}^{k})  \, d\tau \\
&=  L_N(w_{1:N}^k) + \nabla L_N(w_{1:N}^k)^\top(w_{1:N}^{k+1}-w_{1:N}^{k}) + \int_{0}^{1} (\nabla L_N(w_{1:N}^k + \tau(w_{1:N}^{k+1}-w_{1:N}^{k}))-\nabla L_N(w_{1:N}^k))^\top(w_{1:N}^{k+1}-w_{1:N}^k) \, d\tau \\
&\leq  L_N(w_{1:N}^k) + \nabla L_N(w_{1:N}^k)^\top(w_{1:N}^{k+1}-w_{1:N}^{k}) + \int_{0}^{1} \left\Vert \nabla L_N(w_{1:N}^k + \tau(w_{1:N}^{k+1}-w_{1:N}^{k}))-\nabla L_N(w_{1:N}^k)\right\Vert_2 \left\Vert w_{1:N}^{k+1}-w_{1:N}^k \right\Vert_2 \, d\tau.
\end{align}
Under Assumption 3, we obtain
\begin{align} 
L_N(w_{1:N}^{k+1}) & \leq  L_N(w_{1:N}^k) + \nabla L_N(w_{1:N}^k)^\top(w_{1:N}^{k+1}-w_{1:N}^{k}) + \int_{0}^{1} \lambda\left\Vert \tau(w_{1:N}^{k+1}-w_{1:N}^k) \right\Vert_2 \left\Vert w_{1:N}^{k+1}-w_{1:N}^k \right\Vert_2\, d\tau \\
& = L_N(w_{1:N}^k) + \nabla L_N(w_{1:N}^k)^\top (w_{1:N}^{k+1}-w_{1:N}^k) +\frac{\lambda}{2}\left\Vert w_{1:N}^{k+1} -w_{1:N}^k\right\Vert^2_2.
\end{align}
By the SGD update rule (\ref{eq_sgd_cnn}) and (\ref{eq_sgd_rnn}), the above inequality can be written as
\begin{align}  
L_N(w_{1:N}^{k+1}) - L_N(w_{1:N}^k) & \leq  -\alpha_k \nabla L_N(w_{1:N}^k)^\top 
\nabla_w L(w^k, \xi_k)
+ \frac{1}{2}\alpha_k^2 \lambda \left\Vert \nabla_w L(w^k, \xi_k)  \right\Vert ^2_2,
\end{align}
which, under Assumption 1, becomes
\begin{align}  
L_N(w_{1:N}^{k+1}) - L_N(w_{1:N}^k) & \leq  -\alpha_k \sigma \left\Vert \nabla L_N(w_{1:N}^k) \right\Vert_2^2 + \frac{1}{2}\alpha_k^2 \lambda \left\Vert \nabla_w L(w^k, \xi_k) \right\Vert ^2_2.
\end{align}
Taking the expectation with respect to the distribution of $\xi_k$, and noting that $w_{1:i}^{k+1}$ but not $w_{1:i}^k$ depends on $\xi_k$, we obtain the desired bound:
\begin{align}   
\E_{\xi_k} [L_N(w_{1:N}^{k+1})] - L_N(w_{1:N}^k)&  \leq  -\alpha_k \sigma \left\Vert \nabla L_N(w_{1:N}^k) \right\Vert_2^2 + \E_{\xi_k} \left[ \frac{1}{2}\alpha_k^2 \lambda \left\Vert \nabla_w L(w^k, \xi_k) \right\Vert ^2_2 \right].
\end{align}
Under Assumption 2, this becomes
\begin{align}   
\E_{\xi_k} [L_N(w_{1:N}^{k+1})] - L_N(w_{1:N}^k) & \leq  -\alpha_k \sigma \left\Vert \nabla L_N(w_{1:N}^k) \right\Vert_2^2 + \frac{1}{2}\alpha_k^2 \lambda M.
\end{align}
Taking the total expectation $\E[L_N(w_{1:N}^{k})] = \E_{\xi_1}\E_{\xi_2} \dotsc \E_{\xi_{k-1}}[L_N(w_{1:N}^{k})]$ yields
\begin{align} 
\E [L_N(w_{1:N}^{k+1})] -\E [ L_N(w_{1:N}^k)] & \leq  -\alpha_k \sigma \E \left[ \left\Vert \nabla L_N(w_{1:N}^k) \right\Vert_2^2 \right] + \frac{1}{2}\alpha_k^2 \lambda M.
\end{align}
If we take summation from $k=0$ to $K-1$, we obtain 
\begin{align} 
\label{eq25} 
\E [L_N(w_{1:N}^{K})] - L_N(w_{1:N}^0) & \leq - \sigma \sum\limits_{k=0}^{K-1} \alpha_k \E \left[ \left\Vert \nabla L_N(w_{1:N}^k) \right\Vert_2^2 \right] + \frac{1}{2}\lambda M \sum\limits_{k=0}^{K-1} \alpha_k^2.
\end{align}
For the optimal solution of $L_N(w_{1:N}^k)$, $w_{1:N}^*$, 
\begin{align} 
\label{eq26}
L_N(w_{1:N}^*)-L_N(w_{1:N}^0) \leq \E [L_N(w_{1:N}^K)]-L_N(w_{1:N}^0).
\end{align}
Combining (\ref{eq25}) and (\ref{eq26}) yields 
\begin{align} 
L_N(w_{1:N}^*)-L_N(w_{1:N}^0) \leq - \sigma \sum\limits_{k=0}^{K-1} \alpha_k \E \left[ \left\Vert \nabla L_N(w_{1:N}^k) \right\Vert_2^2 \right] + \frac{1}{2}\lambda M \sum\limits_{k=0}^{K-1} \alpha_k^2,
\end{align}
which can be rearranged as 
\begin{align}
\frac{1}{A_K}\sum\limits_{k=0}^{K-1} \alpha_k \E \left[ \left\Vert \nabla L_N(w_{1:N}^k) \right\Vert_2^2 \right] & \leq   \frac{L_N(w_{1:N}^0)-L_N(w_{1:N}^*)}{\sigma A_K} + \frac{\lambda M \sum\limits_{k=0}^{K-1} \alpha_k^2 }{2\sigma A_K},
\end{align}
where $A_k = \sum\limits_{k=0}^{K-1} \alpha_k$.
Since $\lim\limits_{K \to \infty} A_K = \sum\limits_{k=0}^{\infty}\alpha_k=\infty$ and $\sum\limits_{k=0}^{\infty}\alpha_k^2 < \infty$, taking the limit on both sides of the above equation yields
\begin{align} 
\label{eq27}
\lim_{K \to \infty} \E \; \bigg[ \frac{1}{A_K}\sum\limits_{k=0}^{K-1} \alpha_k \left\Vert \nabla L_N(w_{1:N}^k) \right\Vert_2^2 \bigg] = 0.
\end{align}
Thus, we have 
\begin{align}
\label{appx2}
\liminf_{K \to \infty} \E \left[ \left\Vert \nabla L_N(w_{1:N}^k) \right\Vert_2^2 \right] = 0.
\end{align}
Let us define $G(w_{1:N}^k) = \left\Vert \nabla L_N(w_{1:N}^k) \right\Vert_2^2 $. Then, $\nabla G(w_{1:N}^k) = 2 \nabla^2L_N(w_{1:N}^k)\nabla L_N(w_{1:N}^k)$. 
$G(w_{1:N}^{k+1})$ can be written as
\begin{align}   
G(w_{1:N}^{k+1}) & = G(w_{1:N}^k) + \int_{0}^{1} \frac{\partial G(w_{1:N}^k + \tau(w_{1:N}^{k+1}-w_{1:N}^{k}))}{\partial \tau} \, d\tau \\
&=  G(w_{1:N}^k) + \int_{0}^{1} \nabla G(w_{1:N}^k + \tau(w_{1:N}^{k+1}-w_{1:N}^{k}))^\top (w_{1:N}^{k+1}-w_{1:N}^{k})  \, d\tau \\
&=  G(w_{1:N}^k) + \nabla G(w_{1:N}^k)^\top(w_{1:N}^{k+1}-w_{1:N}^{k}) + \int_{0}^{1} (\nabla G(w_{1:N}^k + \tau(w_{1:N}^{k+1}-w_{1:N}^{k}))-\nabla G(w_{1:N}^k))^\top(w_{1:N}^{k+1}-w_{1:N}^k) \, d\tau \\
&\leq  G(w_{1:N}^k) + \nabla G(w_{1:N}^k)^\top(w_{1:N}^{k+1}-w_{1:N}^{k}) + \int_{0}^{1} \left\Vert \nabla G(w_{1:N}^k + \tau(w_{1:N}^{k+1}-w_{1:N}^{k}))-\nabla G(w_{1:N}^k)\right\Vert_2 \left\Vert w_{1:N}^{k+1}-w_{1:N}^k \right\Vert_2 \, d\tau.
\end{align}
Let $\lambda_G$ be the Lipschitz constant of $\nabla G(w_{1:N}^k)$. Then, we obtain
\begin{align} 
G(w_{1:N}^{k+1}) & \leq  G(w_{1:N}^k) + \nabla G(w_{1:N}^k)^\top(w_{1:N}^{k+1}-w_{1:N}^{k}) + \int_{0}^{1} \lambda_G\left\Vert \tau(w_{1:N}^{k+1}-w_{1:N}^k) \right\Vert_2 \left\Vert w_{1:N}^{k+1}-w_{1:N}^k \right\Vert_2\, d\tau \\
& = G(w_{1:N}^k) + \nabla G(w_{1:N}^k)^\top (w_{1:N}^{k+1}-w_{1:N}^k) +\frac{\lambda_G}{2}\left\Vert w_{1:N}^{k+1}-w_{1:N}^{k}\right\Vert ^2_2,
\end{align}
which becomes
\begin{align}
G(w_{1:N}^{k+1})-G(w_{1:N}^k) & \leq  \nabla G(w_{1:N}^k)^\top (w_{1:N}^{k+1}-w_{1:N}^k) + \frac{1}{2}\lambda_G\left\Vert w_{1:N}^{k+1}-w_{1:N}^k \right\Vert_2^2.
\end{align}
By the SGD update rule (\ref{eq_sgd_cnn}) and (\ref{eq_sgd_rnn}),
\begin{align}
G(w_{1:N}^{k+1})-G(w_{1:N}^k) & \leq  -\alpha_k \nabla G(w_{1:N}^k)^\top \nabla_w L(w^k, \xi_k)  + \frac{1}{2}\alpha^2_k \lambda_G\left\Vert \nabla_w L(w^k, \xi_k)  \right\Vert_2^2 \\
& \leq  -2 \alpha_k \nabla L_N(w_{1:N}^k)^\top \nabla^2 L_N(w_{1:N}^k)^\top \nabla_w L(w^k, \xi_k) 	+ \frac{1}{2}\alpha^2_k \lambda_G\left\Vert \nabla_w L(w^k, \xi_k)  \right\Vert_2^2.
\end{align}
Under Assumption 1, we have
\begin{align}
G(w_{1:N}^{k+1})-G(w_{1:N}^k) & \leq  -2 \alpha_k \sigma \left\Vert \nabla L_N(w_{1:N}^k) \right\Vert_2^2 \nabla^2 L_N(w_{1:N}^k)^\top	+ \frac{1}{2}\alpha^2_k \lambda_G\left\Vert \nabla_w L(w^k, \xi_k)  \right\Vert_2^2\\
& \leq  2 \alpha_k \sigma \left\Vert \nabla L_N(w_{1:N}^k)\right\Vert^2_2 \left\Vert \nabla^2 L_N(w_{1:N}^k) \right\Vert_2 + \frac{1}{2}\alpha_k^2 \lambda_G \left\Vert \nabla_w L(w^k, \xi_k)  \right\Vert ^2_2.
\end{align}
Taking the expectation with respect to the distribution of $\xi_k$ yields
\begin{align}   
\E_{\xi_k} [G(w_{1:N}^{k+1})] - G(w_{1:N}^k) & \leq  2 \alpha_k \sigma \left\Vert \nabla L_N(w_{1:N}^k)\right\Vert^2_2 \left\Vert \nabla^2 L_N(w_{1:N}^k) \right\Vert_2 + \E_{\xi_k} \left[ \frac{1}{2}\alpha_k^2 \lambda_G \left\Vert \nabla_w L(w^k, \xi_k)  \right\Vert ^2_2 \right]. 
\end{align}
Under Assumptions 2 and 3, 
\begin{align}   
\E_{\xi_k} [G(w_{1:N}^{k+1})] - G(w_{1:N}^k) &\leq 2 \alpha_k \sigma \lambda \left\Vert \nabla L_N(w_{1:N}^k)\right\Vert^2_2 + \frac{1}{2}\alpha_k^2 \lambda_G M.
\end{align}
By taking the total expectation, we obtain
\begin{align} 
\label{appx} 
\E [G(w_{1:N}^{k+1})] -\E [G(w_{1:N}^k)] &\leq 2 \alpha_k \sigma \lambda \E \left[ \left\Vert \nabla L_N(w_{1:N}^k)\right\Vert^2_2 \right]+ \frac{1}{2}\alpha_k^2 \lambda_G M.
\end{align}
From (\ref{eq27}), $\E \left[ \sum\limits_{k=0}^{\infty}\alpha_k \left\Vert \nabla L_N(w_{1:N}^k) \right\Vert^2 \right] < \infty$. In addition, the theorem assumes $\sum\limits_{k=0}^{\infty} \alpha_k^2 < \infty$. Therefore, if we take summation from $k=0$ to $K$ for (\ref{appx}), we obtain 
\begin{align} 
\label{eq44}
\E [G(w_{1:N}^{K+1})] -\E [G(w_{1:N}^0)] < \infty.
\end{align}
Let us define two nondecreasing sequences
\begin{align}  
S^+_K & = \sum_{k=0}^{K}\max\{0,\E[G(w_{1:N}^{k+1})]-\E[G(w_{1:N}^{k})]\}\\
S^-_K & = \sum_{k=0}^{K}\max\{0,\E[G(w_{1:N}^k)]-\E[G(w_{1:N}^{k+1})]\}.
\end{align}
Since $\E [G(w_{1:N}^{K+1})] -\E [G(w_{1:N}^0)] = S^+_K - S^-_K < \infty$ according to (\ref{eq44}), the positive component of $\E [G(w_{1:N}^{K+1})] -\E [G(w_{1:N}^0)]$, i.e., $S_K^+$, satisfies $S^+_K < \infty$.   
Furthermore, 
\begin{align}
\E [G(w_{1:N}^{K+1})] = \E [G(w_{1:N}^0)] + S^+_K - S^-_K \geq 0
\end{align} 
holds for any $K$, and thus $S^-_K$ also converges.
Therefore, $\E [G(w_{1:N}^K)] =E [\left\Vert \nabla L_N(w_{1:N}^K) \right\Vert _2^2] $ converges and, according to (\ref{appx2}), this limit must be zero, i.e., 
\begin{eqnarray}
\lim_{k \to \infty} \E \left[ \left\Vert \nabla L_N(w_{1:N}^k) \right\Vert_2^2 \right] = 0.
\end{eqnarray}


\end{document}